\documentclass{article}
\usepackage[final]{neurips_2021}
\usepackage[utf8]{inputenc} 
\usepackage[T1]{fontenc}    
\usepackage[pagebackref=false,breaklinks=true,colorlinks,bookmarks=false]{hyperref} 
\usepackage{amsthm}
\usepackage{algorithmic}
\usepackage{amsmath}
\usepackage{amssymb}
\usepackage{url}            
\usepackage{booktabs}       
\usepackage{amsfonts}       
\usepackage{nicefrac}       
\usepackage{microtype}      
\usepackage{amsmath}
\usepackage{tikz-dependency}
\usepackage{tikz}
\usepackage{pifont}
\usepackage{tabularx}
\usepackage{multirow}
\usepackage{subcaption}
\usepackage{lscape}
\usepackage{soul}
\usepackage{amssymb}
\usepackage{adjustbox}
\usepackage{amsmath,bm}
\usepackage{wrapfig}
\def\Eg{\emph{E.g}.}
\def\ie{\emph{i.e}.}
\def\eg{\emph{e.g}.}
\def\ie{\emph{i.e}.}

\def\etc{\emph{etc}}
\newcolumntype{L}{>{\centering\arraybackslash}m{3cm}}

\newcommand{\cmark}{\ding{51}}%
\newcommand{\xmark}{\ding{55}}%
\newcommand{\jx}[1]{\textcolor{black}{#1}} 
\def\mU{{\bm{U}}}

\title{Unified Pretraining Framework for Document Understanding}
\author{%
	{Jiuxiang Gu$^1$, \ Jason Kuen$^1$, \ Vlad I. Morariu$^1$, \ Handong Zhao$^1$},\\
	{\textbf{Nikolaos Barmpalios$^2$, \ Rajiv Jain$^1$, \ Ani Nenkova$^1$, \ Tong Sun$^1$}}\\
	$^1$Adobe Research, $^2$Adobe Document Cloud\\
	\texttt{\{jigu,kuen,morariu,hazhao,barmpali,rajijain,nenkova,tsun\}@adobe.com}
}
\begin{document}
\maketitle
\begin{abstract}
	Document intelligence automates the extraction of information from documents and supports many business applications. Recent self-supervised learning methods on large-scale unlabeled document datasets have opened up promising directions towards reducing annotation efforts by training models with self-supervised objectives. However, most of the existing document pretraining methods are still language-dominated. We present UDoc, a new unified pretraining framework for document understanding. UDoc is designed to support most document understanding tasks, extending the Transformer to take multimodal embeddings as input. Each input element is composed of words and visual features from a semantic region of the input document image. An important feature of UDoc is that it learns a generic representation by making use of three self-supervised losses, encouraging the representation to model sentences, learn similarities, and align modalities. Extensive empirical analysis demonstrates that the pretraining procedure learns better joint representations and leads to improvements in downstream tasks.
\end{abstract}
	
	\section{Introduction}
	Document intelligence is a broad research area that includes techniques for information extraction and understanding. Unlike plain-text documents in natural language processing (NLP)~\cite{yang2016hierarchical,goldstein2000multi}, a physical document can be composed of multiple elements: {tables, figures, charts},~\etc. In addition, a document usually includes rich visual information, and can be one of various types of documents (scientific paper, form, resume,~\etc.), with various combinations of multiple elements and layouts.
	Complex content and layout, noisy data, font and style variations make automatic document understanding very challenging. For example, {to understand text-rich} documents such as letters, a system needs to focus almost exclusively on text content, paying attention to a long sequential context, while processing semi-structured documents such as forms requires the system to analyze spatially distributed short words, paying particular attention to the spatial arrangement of the words. Following the success of BERT~\cite{dai2019transformer} on NLP tasks, there has been growing interest in developing {pretraining} methods for document understanding~\cite{xu2020layoutlm,xu2020layoutlmv2,selfdoc2021}. {Pretrained} models have achieved state-of-the-art (SoTA) performance across diverse document understanding tasks~\cite{park2019cord,harley2015icdar}.
	
	Huge training datasets help pretraining models to learn a good representation for downstream tasks. However, we observe three major problems with the current pretraining setup: \textit{(1)} \textit{documents are composed of semantic regions}. Most of the recent document pretraining works follow BERT and split documents into words. However, unlike the sequence-to-sequence learning in NLP, documents have a hierarchical structure (words form sentences, sentences form a semantic region, and semantic regions form a document). Also, the importance of words and sentences are highly context-dependent,~\ie, the same word or sentence may have different importance in a different context. Moreover, current transformer-based document pretraining models {suffer} from input length constraints. {Also, input length becomes a problem for text-rich} documents or multi-page documents. \textit{(2)} \textit{documents are more than words}. The semantic structure of the document is not only determined by the text within it but also the visual features such as table, font size and style, and figure,~\etc. Moreover, the visual {appearance} of the text {within} a block {are often overlooked}. {Most of recent} BERT-based pre-{training} works only take the words as input without considering {multimodal content and} alignment of multimodal information {within} semantic regions. \textit{(3)} \textit{documents have spatial layout}. Visual and layout information is critical for document understanding. Recent works encode spatial information via 2D position encoding and model {spatial relationships} with self-attention, which computes attention weights for long {inputs}~\cite{xu2020layoutlm,xu2020layoutlmv2}. However, for semi-structured documents, such as forms and receipts, words are more related to their local surroundings. This corresponds strongly with human intuition -- when we look at magazines or newspapers, the receptive fields are modulated by our reading order and attention. Based on the above observations, {we ask the following} question: \textit{Can unified document pretraining benefit all of these different kinds of documents?}
	
\begin{figure*}[ht!]
		\begin{center}
			\includegraphics[width=\linewidth]{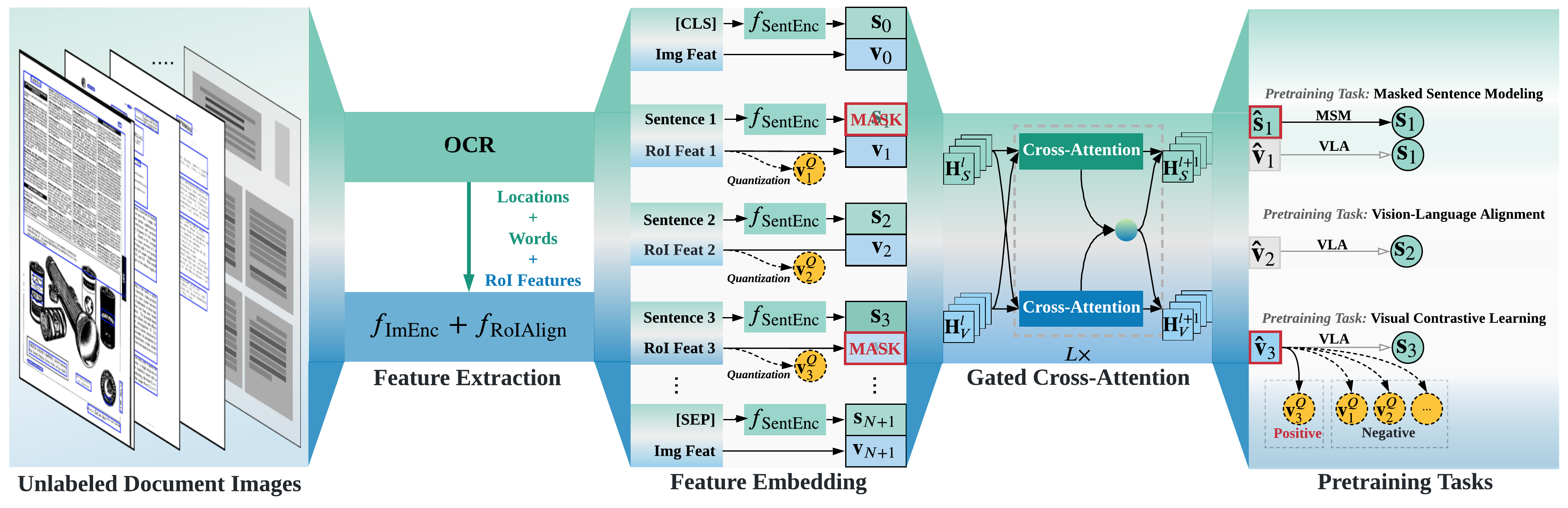}
		\end{center}
		\vspace{-3mm}
		\caption{Overview of the proposed {approach,} UDoc. UDoc first uses a CNN-based visual backbone to learn visual representations. The model {then} extracts the \jx{Region of Interest (RoI)} features with OCR bounding boxes and {generates} a multimodal embedding by combining the textual embedding and position encoding. The transformer-based encoder takes a set of masked multimodal embeddings as input and is pretrained with three pretraining tasks. All the {network} parameters except {those of the} textual encoder are jointly trained {during} both pretraining and fine-tuning phases.}
		\vspace{-5mm}
		\label{fig.pretrain_arch}
	\end{figure*}

	We propose a unified pretraining framework for document understanding, shown in Fig.~\ref{fig.pretrain_arch}. Our model integrates image information in the pretraining stage by taking advantage of the transformer architecture to learn {cross-modal interactions} between visual and textual information. To handle textual information, we {encode sentences using} a hierarchical transformer encoder. The first level of the hierarchical encoder models the formation of the sentences from words. The second level models the formation of the document from sentences. With the help of the hierarchical structure, UDoc learns how words form sentences and how sentences form documents. Meanwhile, it reduces model computation complexity exponentially and increases the number of input words. This also mimics human reading behaviors since the sentence/paragraph is a reasonable unit for people to read and understand{---people} rarely check the interactions between arbitrary words across different {regions} in order to understand {an} article. {Convolution} has been very successful in the extraction of local features that encode visual and spatial information~\cite{gu2018recent}, so we use convolution layers as a more efficient complement to self-attention for addressing local intra-region dependencies in a document image. Meanwhile, self-attention uses all input tokens to generate attention weights for capturing global dependencies. Thus, we combine convolution with self-attention to form a mixed attention mechanism that combines the advantages of the two operations.
	
	We depart from previous vision-language pretraining~\cite{lu2019vilbert,gu2020self} by extracting both the textual and visual features for each semantic region. We propose a novel gated cross-attentional transformer that enables information exchange between modalities. A visually-rich region (figure, chart,~\etc) may have stronger visual information than textual information.
	Instead of treating outputs from both modalities identically, we design a gating mechanism that can dynamically control the influence of textual and visual features. This approach enables cross-modal connections and allows for variable highlight the relevant information in visual and textual modality and enables cross-modal connections. During pretraining, the CNN-based visual backbone and multi-layer gated cross-attention encoder are jointly trained in both pretraining and fine-tuning phases.

	Our contributions are summarized as follows:
	(1) We introduce UDoc, a powerful pretraining framework for document understanding. UDoc is capable of learning contextual textual and visual information and cross-modal {correlations within} a single framework, which leads to better performance.
	(2) We present Masked Sentence Modeling for language modeling, Visual Contrastive Learning for vision modeling, and Vision-Language Alignment for pretraining.
	(3) We present extensive experiments and analyses to validate the effectiveness of the proposed UDoc.
	Extensive experiments and analysis 
	provide useful insights on the effectiveness of {the} pretraining tasks and show outstanding performance on {various} downstream tasks.
	
	\section{Related Work}
	Self-supervised learning has shown great success in producing generic {representations that learn from large-scale unlabeled corpora}~\cite{dai2019transformer}. Like the development of pretraining in computer vision~\cite{deng2009imagenet} and NLP~\cite{dai2019transformer}, there has been a surging interest in self-supervised learning for Vision-Language (VL) tasks~\cite{lu2019vilbert,su2019vl,li2019unicodervl,gu2020self}. {Transformers~\cite{dai2019transformer} are the key technology that enables learning} contextualized representations from large-scale unlabeled training data. \jx{The unique characteristics of document images (spatial layout and multiple elements) distinguish document image pretraining from pretraining works in NLP and VL domains. In the NLP domain, the inputs are pure texts without spatial layouts (bounding boxes). In the VL domain, the inputs are the visual objects and captions. While for document images, the input elements are spatially distributed, and the visual and textual information co-occur within the semantic regions.}
	
	Several recent works have explored pretraining on document images~\cite{xu2020layoutlm,xu2020layoutlmv2,patil2020read}.
	LayoutLM~\cite{xu2020layoutlm} {extends} BERT to learn contextualized word representations for document images through multi-task learning. It takes a sequence of OCR words as input during pretraining and {incorporates the} 2D position embedding {as input} for each token. However, {LayoutLM} only considers textual information during pretraining without modeling the alignment between visual and textual information{--visual information is only incorporated into the model during the fine-tuning stage}. {The most} recent version, LayoutLMv2~\cite{xu2020layoutlmv2}, {improves on this by incorporating} the image encoder into pretraining and jointly {training} the image encoder along with the BERT model. {LayoutLMv2} splits the document image into several parts and concatenates the visual embeddings and text embeddings {into a single} sequence. Apart from \jx{masked language learning (MLM)}, {LayoutLMv2} also considers image-text alignment and image-text matching during pretraining. The most related work to ours is SelfDoc~\cite{selfdoc2021}, which proposes a multimodal document pretraining framework. 
	It first extracts the document object proposals from pre-trained Faster R-CNN~\cite{girshick2015fast} and then applies OCR for each proposal to get the words. It takes the pre-extracted RoI features and sentence embeddings as input, and models the perform learning over the textual and visual information using the cross-modality encoder.
	
	There is a noticeable difference between our {proposed method, UDoc,} and other concurrent works in document \jx{image} pretraining.
	UDoc is a multimodal end-to-end pretraining framework for document images. Unlike the fixed document object detector in~\cite{selfdoc2021}, {the} parameters of {the} image encoder with RoI align, which derive the visual features for semantic regions, are also updated in UDoc. 
	{In contrast to} ~\cite{xu2020layoutlmv2}, our visual features come from the semantic regions instead of {splitting} the image into fixed regions.
	\jx{Like the object-level semantic elements in natural images, for document 
		{images}, we represent the typical document layout elements such as paragraph, title, figure, and table as semantic regions.}
	Moreover, to learn the contextualized visual representations, UDoc masks visual information in the latent space and learns contextualized representations by solving a contrastive learning task defined over a quantization of the latent visual embeddings.
	
	\section{Method}
	\subsection{Model Architecture}
	Fig.~\ref{fig.pretrain_arch} {illustrates} {our} {approach, UDoc, which} consists of four {components}: feature extraction, feature embedding, multi-layer gated cross-attention encoder, and pretraining tasks. Given a document image {and the locations of document elements (sentence or RoI)}, UDoc takes {image regions} and words that correspond to each document elements as {inputs,} 
	and {extracts} their respective embeddings through a visual feature extractor and a sentence encoder. These embeddings are then fed into a transformer-based encoder to learn the cross-modal contextualized embeddings that integrate both visual features and textual features.
	
	In the \textit{feature extraction} step, we first employ an off-the-shelf OCR tool~\cite{easyocr} to extract text from a document image $\mathbf{I}$, where the words are grouped into sentences $\mathcal{S}=\{s_1,\ldots,s_N\}$ whose corresponding bounding boxes are $\mathcal{P}=\{p_1,\ldots,p_N\}$. For each sentence bounding box $p_i$, we use a ConvNet-based backbone $f_{\text{ImEnc}}$ and RoI Align \cite{he2017mask} $f_{\text{RoIAlign}}$ to extract the pooled RoI features $\bm{v}_i$.
	To obtain a \textit{feature embedding}, we extract the sentence embedding $\bm{s}_i$ for each sentence $s_i$ via a pretrained sentence encoder $f_{\text{SentEnc}}$. Each region's RoI feature $\bm{v}_i$ is discretized into a finite set of visual representations $\bm{v}_i^Q \in \mathbf{V}^Q$ via product quantization~\cite{jegou2010product}. The multi-layer \textit{Gated Cross-Attention} encoder takes the position information, masked visual features $\mathbf{\tilde{V}}$ and masked textual features $\mathbf{\tilde{S}}$ as inputs, and then {it} generates the contextualized multimodal representations {($\mathbf{H}_V^{l}$ and $\mathbf{H}_S^{l}$, $l\in [1,L]$)} {and outputs the predicted features ($\mathbf{\hat{V}}$ and $\mathbf{\hat{S}}$)}, where $L$ is the number of stacked {transformer} blocks.
	
	More formally, the pretraining procedure can be decomposed into the following steps:
	\begin{align}\label{eq:full_pipeline}
		\mathbf{I} \xrightarrow[]{\text{OCR}}
		\binom{\mathcal{P}}{\mathcal{S}}
		\xrightarrow[f_{\text{SentEnc}}]{f_{\text{ImEnc}}+f_{\text{RoIAlign}}}
		\binom{\mathbf{V}, \mathbf{V}^Q}{\mathbf{S}}
		\xrightarrow[f_{\text{Mask}}]{}
		\binom{\mathbf{\tilde{V}}}{\mathbf{\tilde{S}}}
		\xrightarrow[]{}
		\binom{\mathbf{H}_V^{l}}{\mathbf{H}_S^{l}}
		\xrightarrow[]{}
		\binom{\mathbf{\hat{V}}}{\mathbf{\hat{S}}}
		\xrightarrow[]{}
		\mathcal{L}_{\text{Pretraining}}
	\end{align}
	where $f_\text{Mask}$ denotes the masking function that randomly mask{s} RoI features and sentence embeddings with {the respective} probabilities $p_{\text{Mask}}^v$ and $p_{\text{Mask}}^s$.
	{$\mathcal{L}_{\text{Pretraining}}$ is composed of  three pretraining tasks: {Masked Sentence Modeling (MSM), Visual Contrastive Learning (VCL), and Vision-Language Alignment (VLA)}.}
	{Next, we provide details mentioned in Eq.~\ref{eq:full_pipeline}.}
	
	\paragraph{Feature Extraction and Embedding.}
	Formally, a document image $\mathbf{I}\in \mathbb{R}^{W\times H}$ consists of $N$ regions, {where each region's bounding box is characterized by a 6-d vector}, as $p_i=\{\frac{x_{\text{LT}}}{W}, \frac{y_{\text{LT}}}{H}, \frac{x_{\text{RB}}}{W}, \frac{y_{\text{RB}}}{H}, \frac{w}{W},\frac{h}{H}\}$, where $w$ and $h$ are of the width and height the {region}, $W$ and $H$ are the width and height of {$\mathbf{I}$}, {while} $(x_{\text{LT}}, y_{\text{LT}})$ and $(x_{\text{RB}}, y_{\text{RB}})$ denote the coordinate{s} of the top-left and bottom-right corner{s} respectively. The 6-d vector is {mapped} into a high-dimensional representation via {a} linear mapping function.
	
	The visual embedding is the sum of the {mapped RoI feature and position embedding}. {Likewise, textual embedding is the sum of sentence embedding and position embedding}. We {also have different types of segments to distinguish different modalities}.
	The input sequence {to the transformer-based encoder} starts with a special start element ([CLS] and full {visual} feature{s}), {then it is followed by} multimodal elements, and {it} ends with a special ending element ([SEP]+full {visual} feature{s}). For the special elements ({[CLS] and [SEP]}), the corresponding full {visual} features are features extracted {from} the whole input image, by applying {$f_{\text{ImEnc}}$} {to} an RoI covering the whole input image.
	
	\paragraph{Quantization Module.}
	Unlike the {fixed image encoder} in~\cite{selfdoc2021}, we jointly learn the image encoder {in an end-to-end fashion} {alongside the multimodal model.} {{A visual} representation can be learned by predicting the visual
		features of the masked regions, but it is challenging to predict such features {exactly, since they} are unconstrained and of continuous representation.} To {{constrain the representation space of the visual features} and {facilitate} the end-to-end learning of image encoder (see Task \#2 in Sec.~\ref{sec:pre_training_tasks})}, {we follow~\cite{oord2017neural,baevski2020wav2vec} and use} vector quantization to discretize the visual features {$\mathbf{V} = \{\bm{v}_1,\ldots, \bm{v}_N\}$} {in}to a finite set of representations {$\mathbf{V}^Q=\{\bm{v}_1^Q,\ldots, \bm{v}_N^Q\}$}.
	Specifically, we define latent embedding spaces {$\bm{e}\in \mathbb{R}^{C\times E}$}, where $C$ is the number of codebooks, and $E$ is the number of entries for each codebook. For each {$\bm{v}_i$}, we first map it to logits {$\bm{v}^{\ell}_i\in \mathbb{R}^{C\times E}$}, and calculate the probability for the $j$-th codebook entry in $i$-th group as $p_{c,e} = \text{exp}((v^{\ell}_{c,e} + g_e)/\tau)/\sum_{k = 1}^{E} \text{exp}((v^{\ell}_{c,k} + g_k)/\tau)$, where $\tau$ is a non-negative temperature, $g_{1:E}$ are i.i.d samples drawn from  Gumbel(0,1) distribution. During the forward pass, we choose one entry vector from each codebook by {$\bm{\tilde{e}}_i\sim \text{argmax}_e p_{c,e}$} and generate the quantized representation {$\bm{v}^Q_i$} by a concatenation of {$\{{\bm{\tilde{e}}_1,\ldots, \bm{\tilde{e}}_G}\}$} {which is then followed by} a linear transformation. During the backward pass, the gradients are {computed through a} Gumbel-Softmax estimator~\cite{jang2016categorical}.
	
	\paragraph{Gated Cross-Attention.}
	To model the {interactions among} multimodal inputs, we introduce {a} multimodal transformer {with gated cross-attention to model the cross-modality relationships.}
	{Let $\mathbf{H}_m^{l+1}$ be output features at the $l$-th layer for one modality $m$, and {let} $n$ be another modality ($m, n \in \{V, S\}$). We obtain the features at $(l+1)$-th layer as}:
	\begin{align}
		\mathbf{H}_m^{l+1} &= f_{\text{LN}} \Big(f_{\text{LN}} \big(\mathbf{H}_m^{l} + f_{\text{Cross-Att}}^l(\mathbf{H}_m^l| \mathbf{H}_n^l)\big) +  f_{\text{FF}}^l \big(f_{\text{LN}}(\mathbf{H}_m^{l} + f_{\text{Cross-Att}}^l(\mathbf{H}_m^l| \mathbf{H}_n^l))\big)\Big)\label{eq:gcatt_vl}
	\end{align}
	{where $f_{\text{LN}}$ denotes layer normalization~\cite{ba2016layer}}.
	{The feed-forward sub-layer $f_{\text{FF}}$ in Eq.~\ref{eq:gcatt_vl} is further composed of two fully-connected sub-layers, both wrapped in residual adds and $f_{\text{LN}}$.}
	
	The core part of Eq.~\ref{eq:gcatt_vl} is the cross-attention $f_{\text{Cross-Att}}(\cdot)$. Given the intermediate representations $\mathbf{H}_m^l$ and $\mathbf{H}_n^l$, the cross-attention output for modality $m$ is computed as:
	\begin{align}
		f_{\text{Cross-Att}} (\mathbf{H}_{m}^l|\mathbf{H}_{n}^l) &= [\text{Cross-Att}^1(\mathbf{H}_{m}^l|\mathbf{H}_{n}^l);\ldots;\text{Cross-Att}^h(\mathbf{H}_{m}^l|\mathbf{H}_{n}^l)]\mU \label{eq:self-attn}\\
		\text{Cross-Att}^i(\mathbf{H}_{m}^l|\mathbf{H}_{n}^l) &= {\text{softmax}\left(f^i_{q}(\mathbf{H}_m^l)f^i_k(\mathbf{H}_{n}^l)^T/\sqrt{d}\right)f^i_v(\mathbf{H}_{n}^l)}
	\end{align}
	{where $f^i_q(\mathbf{H}_m^l)$, $f^i_k(\mathbf{H}_n^l)$, and $f^i_v(\mathbf{H}_n^l)$ are the {\textit{query}}, {\textit{key}}, and {\textit{value}} calculated by linear mapping layers for the $i$-th head. $d$ is the model dimension, $h$ is the number of heads, and $\mU$ is the weight matrix that combines the outputs of the heads.}
	
	Considering the {substantial diversity} of document images {and {the different information needs of differing} document types}, we use a gating mechanism~\cite{hu2018squeeze} to {dynamically weight} the outputs {of {the} visual and textual branches}. Specifically, {we feed the concatenated the visual and textual features to a non-linear network} $f_{\text{Gate}}([\mathbf{H}_m^{l+1} ;\mathbf{H}_n^{l+1} ])$, {which generates the modality-specific attention weights} $\alpha_m^l$ and $\alpha_n^l$, and returns the weights separately to {their respective modality-specific branches to perform element-wise products. We multiply the features for modality $m$ with its modality-specific attention weight, and compute the {updated} feature as: $\mathbf{H}_m^{l+1} = \mathbf{H}_m^{l+1} (1+ \alpha_m^l)$, same that for modality $n$.}
	
	\subsection{Training Tasks and Objectives}\label{sec:pre_training_tasks}
	The full pretraining objective of UDoc (right block in Fig.~\ref{fig.pretrain_arch})  is defined as: $\mathcal{L}_{\text{Pretraining}} = \mathcal{L}_{\text{MSM}} + \mathcal{L}_{\text{VCL}} + \mathcal{L}_{\text{VLA}}$. {In the rest of this section}, we describe each task in detail.
	
	\paragraph{Task \#1$\colon$ Masked Sentence Modeling.}
	This task is similar to the \jx{MLM} task utilized in BERT. The key difference is that we mask sentence{s} instead of tokens. During pretraining, each sentence and RoI {of} the input document is randomly {and independently} masked. For the masked sentence, its token is replaced with a special sentence of [MASK]. The model is trained to predict the masked sentence feature, based on the unmasked words and the visual features. The goal is to predict {the} masked sentence embedding{s} based on the {contextual information from} {the} surrounding sentences and image regions, by minimizing the smooth {L1} loss \cite{girshick2015fast}:
	\begin{align}
		\mathcal{L}_{\text{MSM}}(\Theta)= \sum_i \text{smooth}_{L_1} (\bm{s}_i - f_{\text{UDoc}}(\bm{s}_{i}|\bm{s}_{\setminus  i}, \mathbf{\tilde{V}}))
		\label{eq:msm}
	\end{align}
	{where $\Theta$ is the trainable parameters and $f_{\text{UDoc}}(.)$ outputs the unmasked textual feature, {$\bm{s}_{\setminus  i}$ is the surrounding features for the $i$-th input}, $\mathbf{\tilde{V}}$ are the image features with random masking.}
	
	\paragraph{Task \#2$\colon$ Visual Contrastive Learning.}
	We learn {visual feature} representations by solving a visual contrastive learning task which requires {estimating} the true quantized latent RoI representation. Given {a} {prediction} {$\bm{\hat{v}}_i\in \bm{\hat{V}}$} for the masked RoI {$\bm{\tilde{v}}_i\in\mathbf{\tilde{V}}$}, the model needs to {estimate} the positive quantized representation {$\bm{v}^Q_i$} in a set of quantized candidate representations {$\mathbf{V}^Q$}.
	{Good representations are learned} by maximizing {the} agreement between output representation and quantized representation {of the same RoIs} as follows:
	\begin{equation}
		\mathcal{L}_{\text{VCL}}(\Theta) = -\sum_{\bm{\tilde{v}}_i\in\mathbf{\tilde{V}}}\Big(\log \frac{\exp(\text{sim}(\bm{{\hat{v}}}_{i}, \bm{v}^Q_i)/\kappa)}{\sum_{\bm{v}^Q_j} \exp(\text{sim}(\bm{{\hat{v}}}_i, \bm{v}^Q_j)/\kappa)}\Big) + \lambda  \frac{1}{CE} \sum_{c=1}^{C} \sum_{e=1}^{E} p_{c,e} \log{p_{c,e}}
		\label{eq:loss}
	\end{equation}
	where $\text{sim}(\cdot, \cdot)$ {computes the} cosine similarity between two vectors, {$\lambda$} is a hyperparameter, and $\kappa$ is a temperature scalar. The second term encourages the model to use the codebook entries {more} equally. 
	
	\paragraph{Task \#3$\colon$ Vision-Language Alignment.}
	To enforce the alignment among different modalities,  we explicitly encourage alignment between words and
	image regions via similarity-preserving knowledge distillation~\cite{tung2019similarity}. Note that, unlike the text-image alignment in LayoutLMv2~\cite{xu2020layoutlmv2} which split{s} the image into four regions and predict{s} whether the {given} word is covered or not on the image side, we align the image {and text belonging} to the same region. The goal is to minimize the {differences between the pairwise similarities of} sentence embeddings and the {pairwise similarities of image region features}:
	\begin{align}
		\label{eq:vlaloss}
		\mathcal{L}_{\text{VLA}}(\Theta) = \frac{1}{N\times N}   ||f_{\text{Norm}}({\mathbf{S}\cdot \mathbf{S}^{\top}}) - f_{\text{Norm}}(\mathbf{H}_V^L\cdot \mathbf{H}_V^{L\top})||^2_F
	\end{align}
	where {$\mathbf{S}$ is the unmasked input sentence embeddings, $\mathbf{H}_V^L$ is the mapped visual representations {of} the final layer, $||\cdot||_F$ is the Frobenius norm, and $f_{\text{Norm}}$ performs L2 normalization.}
	
	\section{Experiment}
	\subsection{Pretraining UDoc}
	\paragraph{Pretraining corpus.}
	We build our pretraining {corpus} based on IIT-CDIP Test Collection 1.0~\cite{10.1145/1148170.1148307}, which contains more than 11M scanned document images. To differentiate pretraining from finetuning, we filter out the document images of RVL-CDIP~\cite{harley2015icdar} from IIT-CDIP since it is a subset of IIT-CDIP, and sample 1M document images as our pretraining corpus.
	
	\begin{wraptable}{r}{0.5\textwidth}
		\vspace{-4mm}
		\caption{Comparison of {the} datasets used for pretraining and finetuning process. `Box', `Label', and `Text' {indicate the availability of} location, label and text {annotations} for document entit{ies}. `Tag' denotes the document class label {availability}.}
		\vspace{-2mm}
		\centering
		\small
		\setlength{\tabcolsep}{2pt}
		{
			\begin{tabular}{l|c|c|ccc|c}
				\toprule
				Dataset & Type & Size & Box & Label & Text & Tag \\
				\hline
				IIT-CDIP~\cite{10.1145/1148170.1148307} & Misc & 11M &\xmark & \xmark & \cmark & \xmark \\
				\hline
				RVL-CDIP~\cite{harley2015icdar} & Misc & 400K & \xmark & \xmark & \xmark & \cmark \\
				CORD~\cite{park2019cord} & Receipt & 1K &\cmark & \cmark & \cmark & \xmark \\
				FUNSD~\cite{8892998} & Form & 0.2K & \cmark & \cmark & \cmark & \xmark\\
				PubLayNet~\cite{zhong2019publaynet} & Article & 347K & \cmark & \cmark & \xmark & \xmark \\
				\bottomrule
			\end{tabular}
		}
		\label{table:datasets}
		\vspace{-2mm}
	\end{wraptable}
	
\begin{figure*}[t!]
		\vspace{-1mm}
		\begin{center}
			\includegraphics[width=\linewidth]{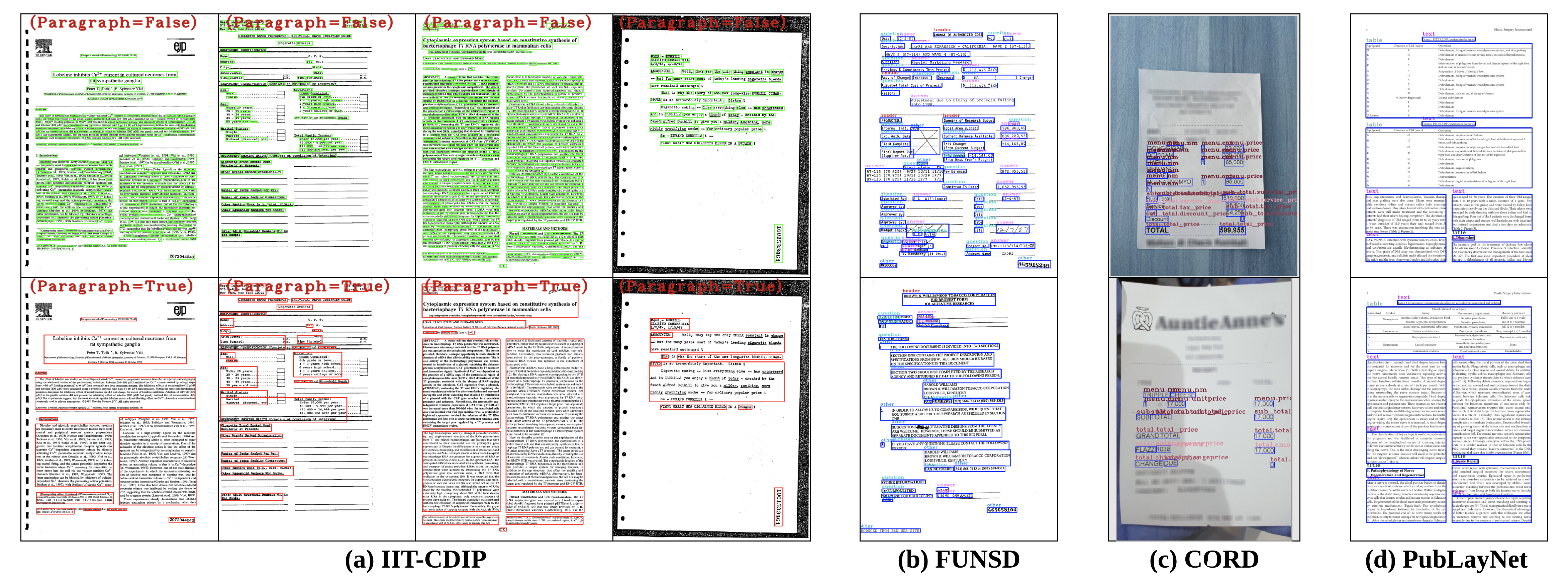}
		\end{center}
		\vspace{-4mm}
		\caption{{Document image samples. The boxes in red/green are OCR bounding boxes {obtained} with/without paragraph mode, {while the} boxes in blue are officially-provided bounding boxes}.}
		\vspace{-4mm}
		\label{fig:dataset_samples}
	\end{figure*}

	Table~\ref{table:datasets} shows the dataset statistics. IIT-CDIP only provides the OCR {texts} in XML format. {We extract} words and their locations by applying EasyOCR~\cite{easyocr} on document images.
	As shown in Fig.~\ref{fig:dataset_samples} (a), EasyOCR provides two kinds of output {modes}: non-paragraph and paragraph. The paragraph mode groups the non-paragraph results {in}to {text} regions.
	{We think document image pretraining should be treated differently than sequence-based pretraining in NLP, since the words in the document (2D) are arranged according to spatial layouts, while the words in NLP corpora are sequential (1D). Considering the special characteristics of documents (complex layout, multi-pages) and the limited input length of BERT models, it is not intuitive to formulate the input at the word level. Hence, we {adopt} the {paragraph-level} outputs as the basic input elements since {textual} regions provide semantically {more} meaningful information than independent words.}
	
	\begin{wrapfigure}{r}{0.5\textwidth}
		\vspace{-4mm}
		\centering
		\small
		\begin{center}
			\includegraphics[width=\linewidth]{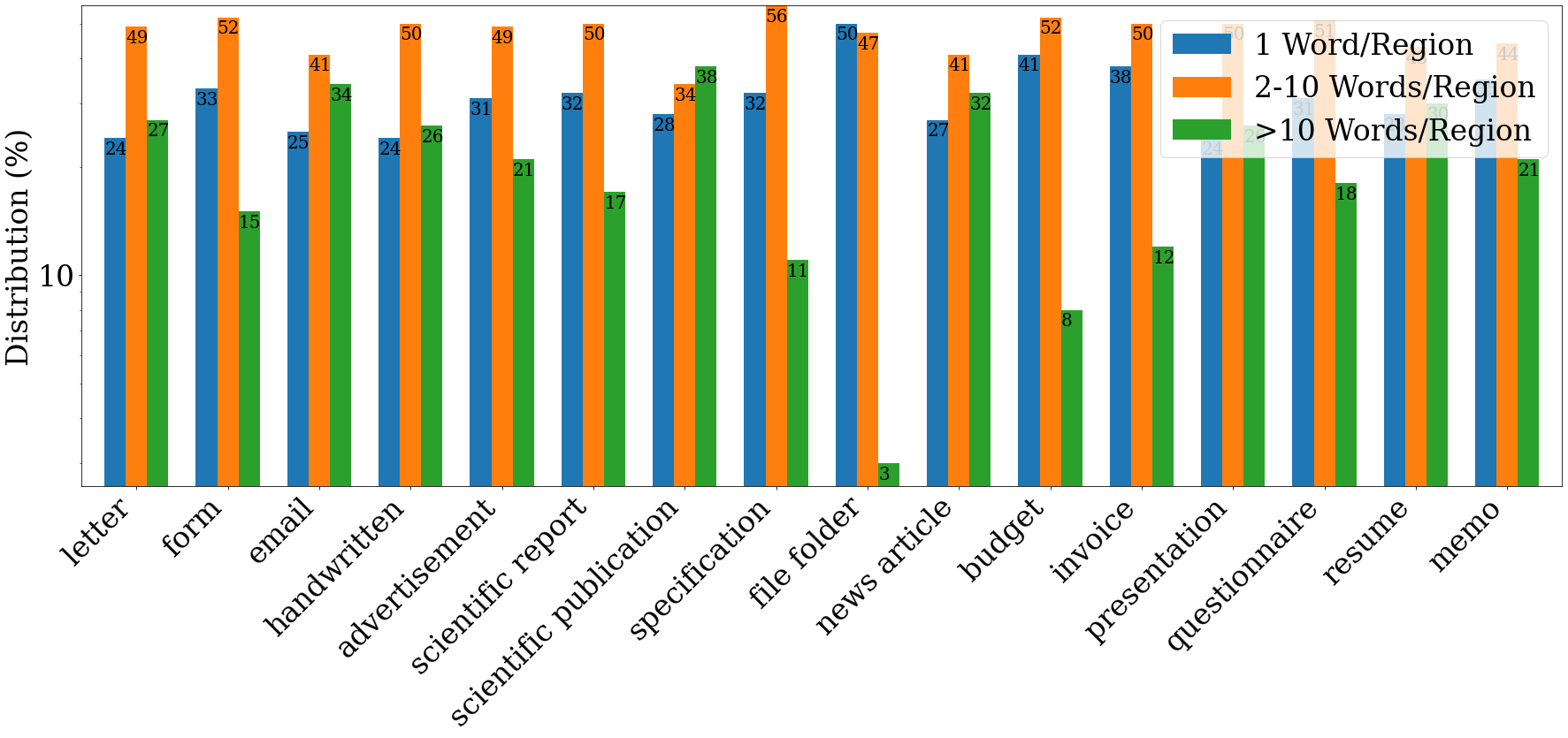}
			\vspace{-3mm}
			\caption{Distribution of words per region on RVL-CDIP according to the categories.}\label{fig:dist_wordregion}
		\end{center}
		\vspace{-7mm}
	\end{wrapfigure}
	\jx{There are some advantages to our design}: (1) {the} region-level design {hierarchically encodes document elements} {and this facilitates the modeling of latent relationships at the region level which has higher-level semantics than the word level.}
	(2) {the hierarchical encoding} {also} overcome{s} the input size limitation of {word-level} BERT-based models~\cite{xu2020layoutlm,xu2020layoutlmv2}.
	\jx{Fig.~\ref{fig:dist_wordregion} shows the distribution of words per region on RVL-CDIP. It can be seen that even though we consider region-level input, for some semi-structured documents, single-words dominate the inputs; this somehow forces UDoc to {pay attention to} word-level inputs. Unlike MLM that predicts the masked word, UDoc predicts the textual embedding of the masked input with MSM.}
	
	\paragraph{Pretraining setting.}
	We initialize the sentence encoder $f_{\text{SentEnc}}$ with BERT-NLI-STSb-base~\cite{reimers-2019-sentence-bert} pretrained for NLI~\cite{williams2017broad} and STS-B~\cite{cer2017semeval}.
	The ResNet-50 backbone in the image encoder is pretrained on the PubLayNet training set~\cite{zhong2019publaynet}.
	All the parameters (except $f_{\text{SentEnc}}$ and $f_{\text{ImEnc}}$) are randomly initialized. During pretraining, we freeze the parameter{s} of $f_{\text{SentEnc}}$ and jointly train the {visual encoder and multi-modal UDoc} model {in an end-to-end fashion}. Such {an} end-to-end training {allows the ConvNet and Transformer to realize their full potentials in spatial and sequence modeling for pretraining.}
	UDoc contains 12 layers {of} gated cross-attention transformer blocks. We set the hidden size to 768 and the number of heads to 12, {the maximum number of regions $N$ to 64, and the maximum input sequence length for $f_{\text{SentEnc}}$} to 512.
	The pretraining is conducted on 8 NVIDIA Tesla V100 32GB GPUs with a batch size of 64. {It is trained with} Adam optimizer \cite{kingma2014adam}, with an initial learning rate of 10$^{-5}$, weight decay of 10$^{-4}$, and {learning rate warmup in the} first 20\% {iterations.}

	To {learn {a} useful} multimodal representation, {random} masking {is applied to both textual and visual inputs}. {For} MSM, we set the mask probability $p_{\text{Mask}}^s$ for input sentences to 15\%. 80\% among the masked sentences are replaced by special sentence [CLS, MASK, SEP], {while} 10\% sentences are replaced by random sentence{s} sampled from other documents, and 10\% remains {unchanged}. {For} VCL, {the $\lambda$ is set to 0.1, $\kappa$ is set to 0.1}, the mask probability $p_{\text{Mask}}^v$ is set to 7.5\% {and the masked} RoI features are {filled} with zeros. {The temperature $\tau$ is annealed from 2.0 to 0.5 by a factor of of 0.999995 at every iteration.} {We select the pretraining checkpoint with the lowest $\mathcal{L}_{\text{Pretraining}}$ for finetuning stage}.
	
	\subsection{Finetuning Tasks}
	\paragraph{Form Understanding.}
	Form understanding requires the model to predict the label for each semantic entity. We use FUNSD~\cite{8892998} as the evaluation dataset. It contains 149/50 training/testing images. Fig.~\ref{fig:dataset_samples} (b) shows {a sample} from FUNSD. Each semantic entity comprises a list of words, a label, and a bounding box. {The officially-provided OCR texts and bounding boxes} are used during training and testing. We take the semantic entities as input and feed the concatenated visual and textual output representation{s} to a classifier. We apply cross-entropy loss {for finetuning}. The model is finetuned for 100 epochs with a learning rate of 10$^{-5}$ and batch size of 16. All the parameters except $f_{\text{SentEnc}}$ are {trained}. {One of} \textit{question}, \textit{answer}, \textit{header} or \textit{other} {is} {predicted} for each semantic entity. We use entity-level F1 score as the evaluation metric.
	
	\paragraph{Receipt Understanding.}
	Receipt understanding requires the model {to} recognize a list of text lines with bounding boxes.
	{The performance on this task is evaluated} on CORD~\cite{park2019cord} dataset. The official data contains 800/100/100 receipts for training/validation/testing. {The receipts are} labeled with 30 types of entities under 4 categories: \textit{company}, \textit{date}, \textit{address}, and \textit{total}. Like FUNSD, we feed the concatenated visual and textual output representation{s} to the classifier. The model is finetuned for 200 epochs with a batch size of 16 and a learning rate of 10$^{-5}$. The evaluation metric is entity-level F1 score.
	
	\paragraph{Document Classification.}
	Document classification {involves predicting} the category for each document image.
	We use RVL-CDIP~\cite{harley2015icdar} as the target dataset. It consists of 320K/40K/40K training/validation/testing images {under} {16 categories}. The OCR words and bounding boxes are extracted by EacyOCR. To fine-tune UDoc on RVL-CDIP, we compute the overall representation as an element-wise product between {the} visual and textual representations {averaged from all sentences/regions}, and learn a classifier {on top of the overall representation} with cross-entropy loss. 
	We fine-tune the model for 30 epochs with a batch size of 64 and a learning rate of 10$^{-5}$. Classification accuracy over 16 {categories} is used to measure model performance.
	
	\paragraph{Document {Object} Detection.}
	Document object detection {involves decomposing} a document image into semantic units. We evaluate the effectiveness of {our} pretrained visual backbone on PubLayNet~\cite{zhong2019publaynet}. As shown Fig.~\ref{fig:dataset_samples} (d), the documents in PubLayNet are scientific articles. PubLayNet consists of 336K/11K training/validation images with six category {labels} (\textit{text}, \textit{title}, \textit{list}, \textit{figure}, and \textit{table}).
	We train Faster-RCNN (F-RCNN) using Detectron2~\cite{wu2019detectron2} and initialize the visual backbone with the pretrained ResNet-50 from UDoc.
	The model is trained for 180k iterations with a base learning rate of 0.01 and a batch size of 8. Mean average precision (MAP) @ intersection over union (IOU) [0.50:0.95] of bounding boxes is used to measure the performance.
	
	\subsection{Results and Discussion}\label{sec:abliation_study}
	\paragraph{The importance of multimodal learning.}
	To study the effect of multimodal learning, we experiment {in} three different settings (1) Vision only (V): this {setting omits the textual components of} UDoc and adopts multilayer self-attention transformer to learn the visual representation. (2) Language only (L): this setting {omits} the {visual} encoder and keeps {only} the {textual components}. (3) Vision-Language (V+L): this setting considers both vision and language {information}. We first train three settings without pretraining. Table~\ref{tab:ab_study} shows consistent improvement across tasks for V+L over the single-stream baselines (V or L). 
	This demonstrates that our UDoc model is able to learn important visual-linguistic relationships that {benefit} downstream tasks {even without pretraining}.
	
	In Table~\ref{tab:ab_study}, we {find} that {visual} information dominates the performance of document classification, while language information contributes a lot to form understanding and receipt understanding. The results also {indicate} that different document tasks rely on different information. For document entity recognition tasks, language information {is more important} than visual features. As can be seen in Fig.~\ref{fig:dataset_samples} (b) and (c), entity recognition is more {word-oriented}. {On the other hand,} document classification is more focused on global{-level} understanding. {As a result, } visual and layout information contribute a lot to the final prediction {of the document classification model}. This {matches well} with {the innate abilities of humans to distinguish between} {document types} without {fully understand{ing}} the words. \jx{We also observe that gated cross-attention (V+L) achieve{s} {a} better performance than the non-gated version (V+L$^\sharp$), as {its} gating mechanism can learn {to} adaptively determine how much each modality contributes to the output features.}
	
	\paragraph{Effect of pretraining tasks.}
	We analyze the effectiveness of different pretraining settings through ablation studies over FUNSD, CORD$^\dagger$, and RVL-CDIP, {which are} representative document benchmarks. Table~\ref{tab:ab_study} ablates {the} key design choices in pretraining UDoc. Note that, for CORD$^\dagger$, $\dagger$ indicates that we use different splits for the ablation study instead of the official ones. For experimental efficiency, UDoc {models evaluated here} are {trained with} 5 epochs on 300k training corpus. Overall, the pretraining of UDoc {consistently} improves the performance over all three downstream tasks. The improvement {gains vary among} different tasks.
	
	\begin{table}[ht]
		\vspace{-4mm}
		\caption{Experimental results and comparison on FUNSD, CORD$^\dagger$, and RVL-CDIP test set{s}.}\label{tab:ab_study}
		\centering
		\small
		\setlength{\tabcolsep}{0.8pt}{
			\begin{tabular}{c|c|c|c|r|c|c|ccc}
				\toprule
				\multicolumn{7}{c|}{Pretraining} & FUNSD & CORD$^\dagger$ & RVL-CDIP  \\
				\cline{1-10} 
				Enable & \#Data  & Modality & Max \#Words & \#Param. & Tasks & Epoch & F1 & F1 & Accuracy \\
				\hline
				\multirow{4}{*}{\xmark }   & -- & V & --  & 85M & -- &  --& 77.49 & 57.08 & 91.35\\
				& -- &  L & -- & 153M & -- &  --&  78.46 &71.52 &  86.82 \\
				& -- &  V+L$^{\sharp}$ & -- & 255M & -- &  --&  80.60 &95.98 &92.76 \\
				& -- &  V+L & -- & 267M & -- &  --&  83.34 &96.59 &92.93 \\
				\hline
				\multirow{4}{*}{\cmark } 
				& 300K &V+L & $64\times 512$  & 270M & MSM + MVM & 5 &  84.37 & 97.44 & 93.10  \\
				& 300K &V+L & $64\times 512$ & 272M & MSM + VCL &  5 & 86.87  & 98.70 & 93.59 \\
				& 300K &V+L & $64\times 512$ & 272M & MSM + VCL + VLA &  5 & 87.38 & 98.75 & 93.92 \\
				& 300K &V+L & $64\times 512$ & 274M & MSM + VCL + VLA + REL & 5 & 87.20 &98.13 &  93.64\\
				\bottomrule
		\end{tabular}}
		\vspace{-3mm}
	\end{table}
	We {first establish two baselines: MSM+MVM in Table~\ref{tab:ab_study} indicates the combination of masked sentence learning and masked visual feature prediction. Similar to MSM, for MVM, we freeze the visual backbone and perform masked visual feature prediction via RoI-feature regression.}
	MSM+VCL {jointly trains the visual backbone end-to-end} with contrastive learning.
	{As shown in Table~\ref{tab:ab_study}}, MSM+MVM achieves better results than the model without pretraining. {Furthermore,} when combining VCL together with MSM, consistent performance gain{s are observed} across all the benchmarks. Among the three finetuning tasks, the improvements {on} FUNSD and CORD are bigger than {on} RVL-CDIP. {We think} the  {local context modeling capability of the ConvNet-based image encoder brings more benefits to entity recognition, since entities are heavily linked and correlated to their local surroundings.}
	When MSM, VCL, and VLA are jointly trained, we observe {further} performance gain{s} across all the benchmarks.
	\jx{For VCL, instead of sampling the negatives from the same input document, we also {try including} the negative samples from other document images of the same batch. However, we {find} that sampling negatives from the entire batch of document images hurt the performance. This is likely because the negatives from other document images are easy to distinguish from each other.}
	{We also consider the image-text {matching} task (Rel)~\cite{xu2020layoutlmv2}, and combine Rel with MSM+VCL+VLA}. It hurts the performance of all three downstream tasks.
	{We conjecture that the} image-text matching {task} introduce{s} {mismatched pairs of} image and OCR texts as negative examples {that potentially hamper} the training of other tasks.
	
	\paragraph{What if Masked Language Modeling is included?}
	{To {study the feasibility of that}, we consider MLM during pretraining. Since the number of words may be very {large}, we select the tokens by randomly applying a sliding window (window size 128) across all sequenced OCR words. Each word is formulated as a single-word sentence ([CLS] [Token] [SEP]). We randomly mask 15\% of those sampled words ([CLS] [MASK] [SEP]) and concatenate them along with the region-based inputs. During pretraining, we add the word prediction head on top of UDoc and predict the masked words. We conduct finetuning experiments on entity recognition tasks (FUNSD and CORD), and find that such a direct combination hurts the performance: FUNSD: 87.38 (UDoc) vs. 83.76$\downarrow$ (UDoc+MLM), CORD: 98.75 (UDoc) vs. 98.63$\downarrow$ (UDoc+MLM). There are consistent performance drops from adding MLM. One possible reason is that the RoI features extracted by token bounding boxes might not be discriminative enough due to the tiny word-level bounding boxes}.
	
	\begin{figure*}[t!]
		\begin{center}
			\includegraphics[width=\linewidth]{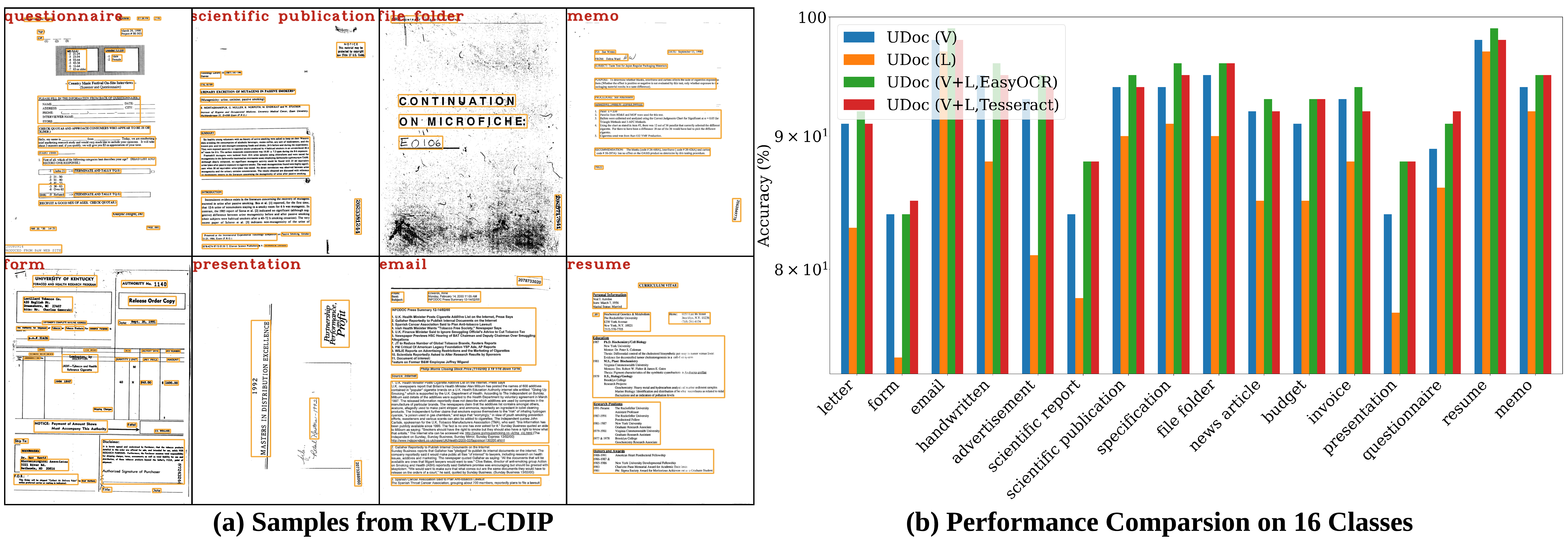}
		\end{center}
		\vspace{-3mm}
		\caption{For (a) we show the samples from RVL-CDIP. The boxes in orange color are grouped OCR bounding boxes. For (b) we plot the accuracies on 16 classes achieved by different models that are represented by different colors in the bar chart.}
		\vspace{-6mm}
		\label{fig.doccls_samples}
	\end{figure*}
	
	\begin{table}[ht]
		\vspace{-4mm}
		\caption{Comparison with state-of-the-art methods. \jx{The symbol $\ddag$ implies using Google OCR engine.}.}
		\centering
		\small
		\setlength{\tabcolsep}{0.7pt}{
			\begin{tabular}{l|c|c|c|c|c|c|ccc}
				\toprule
				\multirow{2}{*}{Method} & \multicolumn{6}{c|}{Pretraining} &  FUNSD & CORD & RVL-CDIP \\
				\cline{2-10} 
				& Source & \#Data  & Scale & Max \#Words & Modality & \#Param. & F1 & F1 & Accuracy \\
				\hline
				$\text{BERT}_{\text{BASE}}$~\cite{xu2020layoutlmv2} & -- & -- &  Word & 512 & L & 110M & 60.26 & 89.68 &  89.81 \\
				$\text{BERT}_{\text{LARGE}}$~\cite{xu2020layoutlmv2} & -- & -- & Word & 512 & L & 340M & 65.63 & 90.25 &  89.92 \\
				$\text{LayoutLM}_{\text{BASE}}$~\cite{xu2020layoutlmv2} & IIT-CDIP & 11M & Word & 512 & L &  113M &  78.66 & 94.72 & 94.42 \\
				$\text{LayoutLM}_{\text{LARGE}}$~\cite{xu2020layoutlmv2} & IIT-CDIP & 11M & Word & 512 & L & 343M & 78.95 & 94.93 & 94.43 \\
				
				$\text{LayoutLMv2}_{\text{BASE}}$~\cite{xu2020layoutlmv2} & IIT-CDIP & 11M & Word & 512 & V+L & 200M & 82.76  & 94.95 & 95.25 \\ 
				$\text{LayoutLMv2}_{\text{LARGE}}$~\cite{xu2020layoutlmv2} & IIT-CDIP& 11M & Word & 512 & V+L & 426M & 84.20 & 96.01 & 95.64 \\
				SelfDoc~\cite{selfdoc2021} & RVL-CDIP & 320K & Region & 50$\times$512 & V+L & -- & 83.36  & -- & 92.81 \\
				SelfDoc+VGG-16~\cite{selfdoc2021} & RVL-CDIP & 320K & Region & 50$\times$512 & V+L & -- & --  & -- & 93.81 \\
				TILT-Base~\cite{powalski2021going} & RVL-CDIP+ & 1.1M & Word & 512 & V+L & 230M & -- & 95.11 & 95.25 \\
				TILT-Large~\cite{powalski2021going} & RVL-CDIP+ & 1.1M & Word & 512 & V+L & 780M & -- & 96.33 & 95.52 \\
				\hline
				UDoc & IIT-CDIP & 1M & Region & 64$\times$512 & V+L & 272M & 87.96 & 96.64 & 93.96 \\
				UDoc$^{\ast}$ & IIT-CDIP & 1M & Region & 64$\times$512 & V+L & 272M & 87.93 & 96.86 & 95.05$^\ddag$ \\
				\bottomrule
		\end{tabular}}
		\label{tab:stoa}
		\vspace{-4mm}
	\end{table}
	
	\paragraph{Performance Comparison with SoTA.}
	We further pretrain UniDoc on 1M document images with 5 epochs and report the finetuning results in Table~\ref{tab:stoa}.
	{UniDoc} outperforms previous models on the official test set of FUNSD and CORD, by a significantly large margin{, demonstrating that our proposed approach} is highly effective, partially {due to the end-to-end training} of {the} image encoder {that improves} the semantic alignments between images and texts. Note that UniDoc is pretrained on a subset of IIT-CDIP (1M document images), which is {considerably} less than {the} 11M document images used in LayoutLM~\cite{xu2020layoutlm} and LayoutLMv2~\cite{xu2020layoutlmv2}.
	{TILT~\cite{powalski2021going} builds a 1.1M pretraining corpus by combining RVL-CDIP, UCSF Industry Documents Library, and Common Crawl.}
	UniDoc also achieves promising results on document classification. {Note that} {both} LayoutLM v2 and TILT {use} Microsoft OCR, which is a commercial {service} {with a stronger OCR performance than {EasyOCR, which is used in our experiments}.} {We find that OCR plays a key role in document classification performance.} As shown in Fig.~\ref{fig.doccls_samples}, UniDoc performs {the} best {on} {the} `\textit{email}' category {but} worst {on} the `\textit{form}' category.
	\jx{We also report the results with different OCR engines: 93.42 (Tesseract~\cite{10.5555/1288165.1288167}) vs. 93.96 (EasyOCR~\cite{easyocr}) vs. 94.10 (Google OCR~\cite{googleocr}). UniDoc with EasyOCR achieves {a} better performance {than with Tesseract} since EasyOCR is {powered by} an advanced neural network, while Tesseract is based on less sophisticated techniques.}
	\jx{Since different tasks require task-specific input embeddings to perform well, instead of finetuning the sentence encoder during pretraining, we explore unfreezing the sentence encoder during the finetuning stage (named as UniDoc$^{\ast}$) and report the results in Table~\ref{tab:stoa}. Unsurprisingly, we see performance improvements on several downstream applications. \Eg, RVL-CDIP: 93.96 (UniDoc) vs. 95.05$\uparrow$ (UniDoc$^{\ast}$). However, this also makes the training more challenging in terms of computational resources and training time.}
	
	\begin{wraptable}{r}{0.6\textwidth}
		\vspace{-4mm}
		\small
		\caption{MAP @ IOU [0.50:0.95] of the document detection models on PubLayNet dev set.}
		\vspace{-2mm}
		\setlength{\tabcolsep}{1.5pt}
		{
			\begin{tabular}{l|cccccc}
				\toprule
				Method &Text & Title & List & Table & Figure& mAP \\
				\hline
				F-RCNN (ResNet-101)~\cite{zhong2019publaynet} & 91.0 & 82.6 & 88.3 & 95.4 & 93.7 & 90.0 \\
				M-RCNN (ResNet-101)~\cite{zhong2019publaynet} & 91.6 & 84.0 & 88.6 & 96.0 & 94.9 & 90.7 \\
				\hline
				F-RCNN (ResNet-50) &  92.2 & 84.4 & 89.5 & 96.5 & 94.5 & 91.4 \\
				F-RCNN (UDoc, ResNet-50) & 93.9  & 88.5 & 93.7 & 97.3 & 96.4 & 93.9 \\
				\bottomrule
			\end{tabular}
		} 
		\label{table:detection_compare}
		\vspace{-3mm}
	\end{wraptable}
	
	\vspace{1mm}
	\noindent\textbf{Effect of visual backbone.}
	{Additionally,} we apply the trained visual backbone to document {object} detection on PubLayNet. The performance of the F-RCNN on {the validation set} is depicted in Table~\ref{table:detection_compare}. To better {compare}, we {establish} two {F-RCNN models with}: (1) {backbone initialized with ResNet-50 pretrained on ImageNet;} (2) {backbone initialized from UDoc's pretrained visual backbone.} {It can be seen that our pretrained backbone outperforms ImageNet-pretrained backbones.} {By leveraging UDoc}, we can train different variants of the visual backbone and apply them {to document-specific} downstream applications, { without relying on incompatible pretrained backbones from other domains (\eg, natural image).} {Moreover, the visual backbone of UDoc {does not} require any {custom} layers, and thus {any ConvNet architecture can be used in place of ResNet.}}
	
	\section{Conclusion, Limitations, and Future Works}
	{We} develop UDoc, a unified pretraining framework for document understanding.  Our model introduces a novel joint training framework that effectively exploits the visual and textual information during pretraining and finetuning. We evaluate the UDoc comprehensively on three downstream tasks: form understanding, receipt understanding, and document image classification. Extensive empirical analysis demonstrates that the pretraining procedure can take advantage of multimodal inputs. Also, it can effectively aggregate and align visual and textual information of document images with the proxy tasks. This work has a broader impact on document applications. By finetuning the pretrained UDoc on task-specific data, document processing systems can provide better results and reduce {the expensive data annotations costs}. In terms of negative social impact, {the} document images {used for pretraining may} contain sensitive information {and therefore the} models trained on such data may {inappropriately} leak some {private} information. To address the privacy leakage, it is worthwhile to explore the combination of privacy-preserving {learning} and self-supervised learning.
	
	There are {interesting} {short- and long-term} research directions {for} UDoc:
	(1) we freeze the sentence encoder during pretraining and fine-tuning {phases} due to {computational constraints}. {A} better document representation can be learned by jointly training {the} sentence encoder, visual backbone and cross-attention encoder in a {completely} end-to-end fashion.
	(2) {Although} impressive performance has been achieved in document entity recognition tasks such as form and receipt understanding, the classification accuracy {on} semi-structured documents such as forms is still {inferior to that of} rich-text documents. {It is possible to devise} a better method to model the spatial relationship among words.
	(3) {An interesting direction is to} extend UDoc to multipage/multilingual document pretraining. Additionally, there {exist} many text-based labeled document datasets in the NLP domain, such as document summarization.
	Can we transfer the knowledge learned from the text-based document domain to the image-based document domain? \jx{How to unify the pretraining of the pure-text document (1D) and image-based document (2D) in a single framework is also worth to try.}
	Lastly, \jx{the use of different OCR tools is one of the major sources of inconsistency among the existing document pretraining works. It is worthwhile and essential to build {standardized} 
		pretraining document image dataset{s with preprovided OCR results}.} In addition to scanned documents, {using} digital PDF as {part of the} pretraining data is {a direction} worth exploring since it provides rich metadata {which {could be} beneficial for} multimodal learning.
	
	\bibliographystyle{unsrt}
	\bibliography{ref}
\end{document}